\documentclass[letterpaper, 10 pt, conference]{ieeeconf} 
\IEEEoverridecommandlockouts
\usepackage{graphicx} 
\usepackage{epsfig} 
\usepackage{mathptmx} 
\usepackage{times} 
\usepackage{amsmath} 
\usepackage{amssymb}  
\usepackage{url}

\usepackage{booktabs}
\usepackage{physics}
\usepackage{float}
\usepackage{dblfloatfix}
\usepackage{textcomp}
\usepackage{gensymb}
\usepackage{multicol,tabularx,capt-of}
\usepackage{multirow}

\usepackage{color}

\title{\LARGE \bf
Enabling the Sense of Self in a Dual-Arm Robot
}

\author{Ali AlQallaf$^{1}$ and Gerardo Aragon-Camarasa$^{1}$
\thanks{$^{1}$ Computer Vision and Autonomous group, School of Computing Science, University of Glasgow. Email: {\tt\small a.alqallaf.1@research.gla.ac.uk, gerardo.aragoncamarasa@glasgow.ac.uk}}%
}

\begin{document}

\maketitle
\thispagestyle{empty}

\begin{abstract}
While humans are aware of their body and capabilities, robots are not. To address this, we present in this paper a neural network architecture that enables a dual-arm robot to get a sense of itself in an environment. Our approach is inspired by human self-awareness developmental levels and serves as the underlying building block for a robot to achieve awareness of itself while carrying out tasks in an environment. We assume that a robot has to know itself before interacting with the environment in order to be able to support different robotic tasks. Hence, we implemented a neural network architecture to enable a robot to differentiate its limbs from the environment using visual and proprioception sensory inputs. We demonstrate experimentally that a robot can distinguish itself with an accuracy of $88.7\%$ on average in cluttered environmental settings and under confounding input signals. 
\end{abstract}

\section{INTRODUCTION}\label{sec:intro}

When we become self-aware, we can recognise ourselves in any environment. This is possible because we can distinguish our body as a separate entity from the world; allowing us to adapt to different situations and scenarios. Robots, however, lack this capability because they are limited to fixed configurations, engineered to work in constrained environments. Researchers have theorised \cite{vasconez2019human,Kwiatkowskieaau9354,torras2013turing,10.3389/frobt.2018.00088} that an adaptable robot can increase its productivity, and that a self-aware robot can increase the task efficiency over different settings and environments. Hefner et al.~\cite{10.3389/fnbot.2020.00005} has stated that the pathway to reach a similar level of human performance in robotics, a robot requires to have a minimal self and, ultimately, knowledge of self. Current research in self-awareness for robotics \cite{Kwiatkowskieaau9354,sancaktar2019end} has focused on enabling robots and agents to acquire self-awareness by interacting with the environment and task using end-to-end approaches. Self-awareness is, however, learned as a byproduct of the robot interacting with the environment in order to increase its autonomy. 

\begin{figure}[t]
    \centering
    \includegraphics[width=8.5cm]{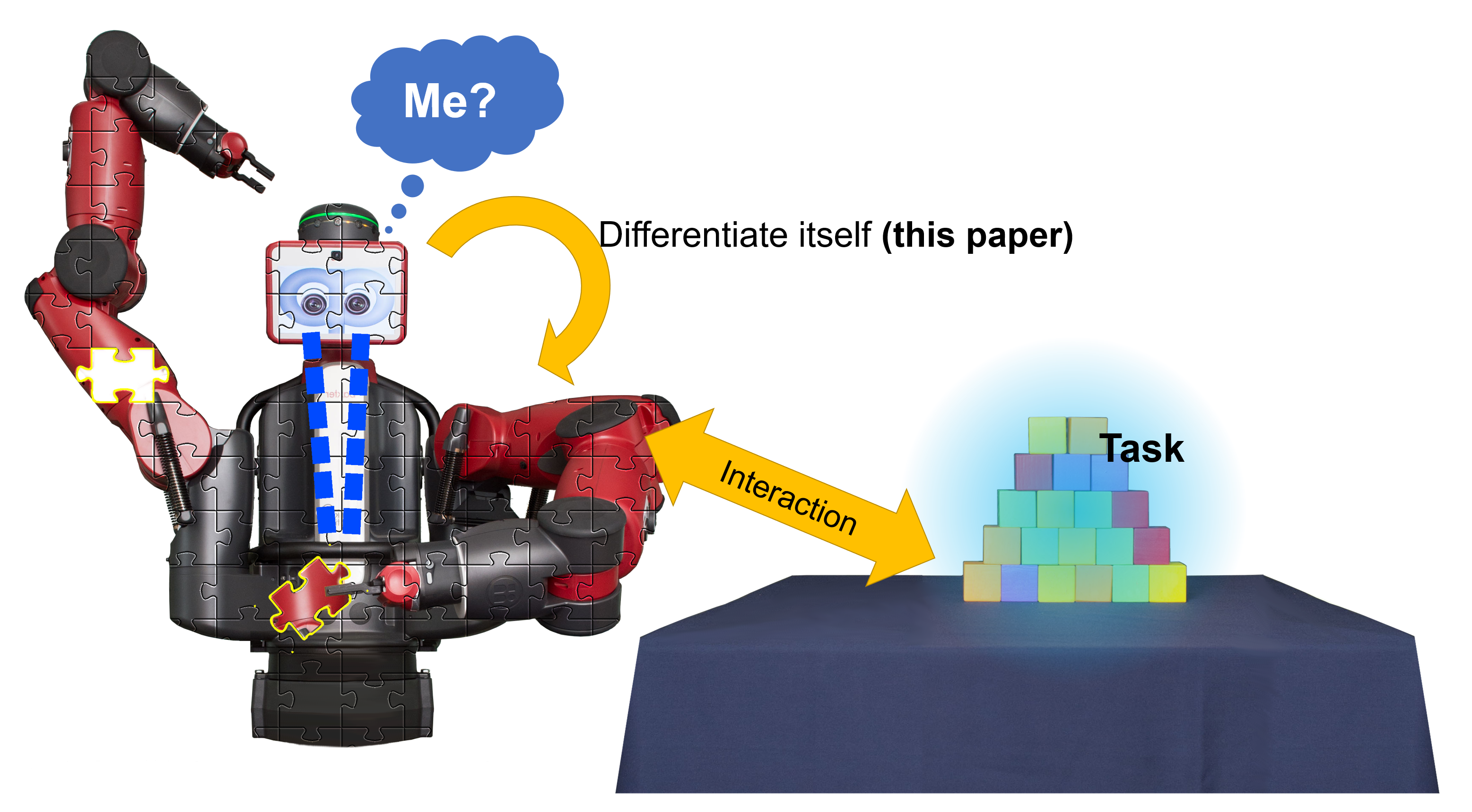}
    \caption{A robot differentiates, recognises and situates itself first with its body, and then interacts with the environment.}
    \label{fig:in-to-out}
\end{figure}

Rochat \cite{rochat2003five} has argued that self-awareness in humans is an incremental process which humans start acquiring it by learning their body and capabilities first, then adapt their self-agency within the environment to fulfil high-level tasks. Rochat~\cite{rochat2003five} has thus proposed five levels of self-awareness, each representing a competence that humans use to learn and adapt to its body and then to environments. In this paper, we propose that a robot starts by learning how to construct a sense of \textit{self}, before interacting and dealing with the environment and objects, as shown in Fig. \ref{fig:in-to-out}. Our approach contrasts to previous and current approaches, where the self is built following a top-down approach via the interaction with the environment ~\cite{10.3389/frobt.2018.00088,Tani:1998:1355-8250:516,6037335,lanillos2020robot}. Hence, we investigate the first basic level of self-awareness which serves as the building block for enabling a robot to become an adaptable and flexible autonomous machine. For this, the robot needs to correlate its visual and internal sensing modalities to initiate the sense of self. We frame this basic level of self-awareness as a binary classification task in which we let the robot to answer whether it can differentiate itself as an entity in an environment with a certain degree of certainty (i.e. certainty is the accuracy of the classification prediction). We evaluate our approach to artificial self-awareness in a dual-arm robot in four different experimental scenarios and carry out an ablation study to investigate whether the robot can differentiate itself with a certain degree of certainty while presented with confounding signals.
Our code and dataset for this paper are available at \url{https://github.com/cvas-ug/towards-sa}.

\begin{figure*}[t]
    \centering
    \includegraphics[width=13cm] {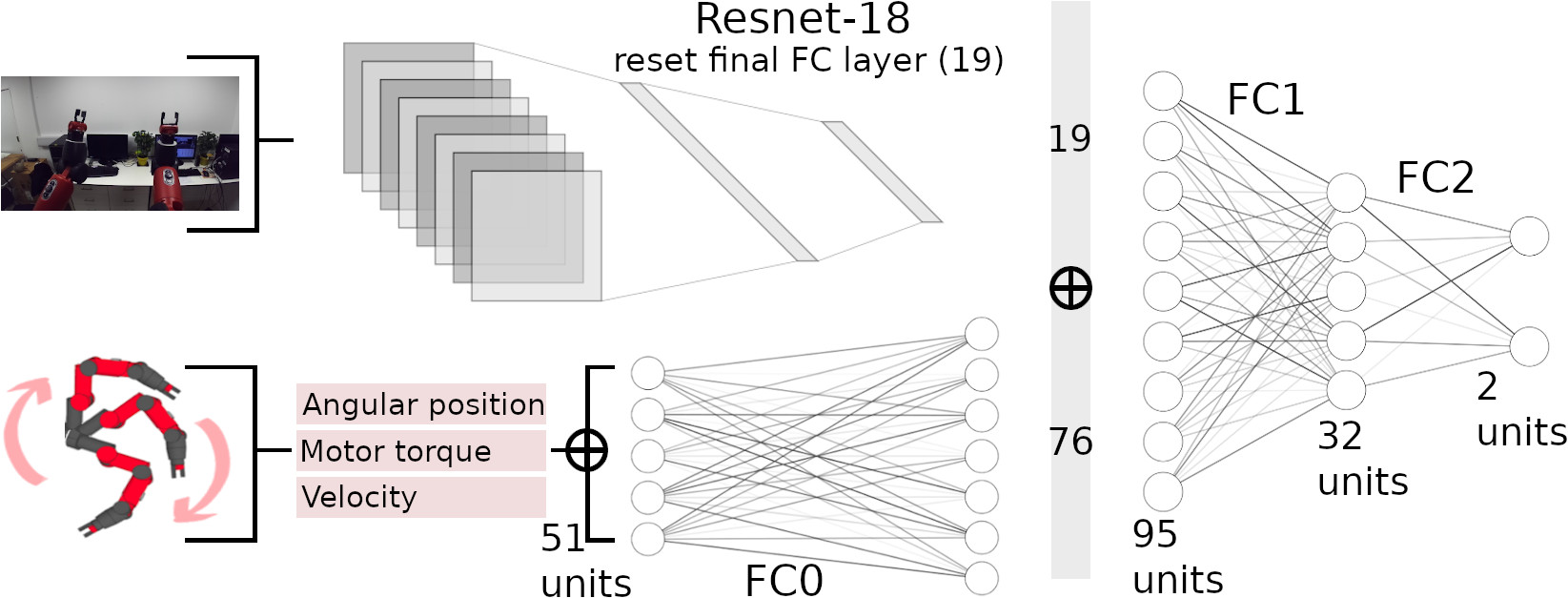}
    \caption{The Level 1 architecture combines vision and proprioception inputs of the robot sensors to predict \textit{self} or \textit{environment}. As shown above, the model process vision and proprioception through two subnetworks (Resnet18 and Linear layer, respectively), then we concatenate the output features from both subnetworks and then passed to 3 fully connected layers to carry out a prediction.}
    \label{fig:selfy_model}
\end{figure*}

 
\section{RELATED WORK}\label{sec:LR}

Rochat~\cite{rochat2003five} has classified self-awareness into five levels, starting from sensing self as a separate entity in the world (Level 1) to self-consciousness (Level 5). Later, Rochat~\cite{10.3389/fpsyg.2019.00414} proposed that \textit{self-unity} (Level 0) is the primary phase of newborns which comprises the initial experience of sensory during the first hours of life, and concluded that self-unity could endow machines to learn about their body within an environment.
The ordering of the five levels of self-awareness is based on their relative complexity and are further divided into implicit (from zero to two) and explicit (from three to five) levels~\cite{rochat2003five, legrain2011distinguishing}. That is, Legrain et al.~\cite{legrain2011distinguishing} have formulated that the implicit self-awareness levels are related to correlating the internal states with the body based on the experience of the self within the environment. The explicit self-awareness levels are those that link the environment to how the environment influences the person. In this paper, we focus on the first self-awareness level and, for completeness, we summarise the implicit levels of self-awareness according to \cite{rochat2003five}:

\begin{itemize}
    \item Level Zero -- "Self-unity". An individual is born with basic multi-sensory and motor control capabilities which they use to learn about itself.
    \item Level One -- "Differentiation". The individual gets a sense that there is something unique in the experience between what is out there and the felt movements to initiate the sense of self.
    \item Level Two -- "Situation". An individual situates within its body by experiencing the relationship between seen movements and body stimulation over time.
\end{itemize}

In robotics, Torras~\cite{torras2013turing} and Chatila et al.~\cite{10.3389/frobt.2018.00088} have stated that there is a need for robots to be capable of handling different environments while showing high adaptability to any environment. However, Agostini et al.~\cite{agostini2011integrating} have argued that robots cannot accommodate all human environments, and hard-coding all possible situations is a challenging task. To mitigate this, researchers~\cite{sancaktar2019end,lanillos2018active} have proposed to learn an awareness model inspired by the free-energy principle~ \cite{friston2010free} in robotics which states that the interactions with the environment are aimed at reducing the internal entropy (i.e. maximising the robot's self-certainty) of an agent. For example, \cite{sancaktar2019end,lanillos2018active} has shown that a robot or its environment might change, and the capability of the robot to adapt to different environments is predicated on the assumption that a robot learns continuously using an active inference model. They thus enabled a robot to adjust its control to the task at hand by minimising the distance between the robot's hand and the target object~\cite{lanillos2018active} or where the robot's hand is to its internal belief~\cite{sancaktar2019end}. However, the authors constrained the robot to have reduced visual perception capabilities in order to simplify the inference task, relying on an observed action within an uncluttered, simple operating environment.

Kwiatkowski et al.~\cite{Kwiatkowskieaau9354} have shown that a robot can model itself without prior knowledge of its structure, and constructs a self-model that can adapt to mechanical changes that occur to the robot. Their work has demonstrated that self-modelling is the conduit to adaptable and resilient robotic systems. However, the proposed self-model architecture learns about the robot's internal mechanical structure, and it is not able to make a distinction of itself as an entity in the environment without being explicitly defined. The basic robot's existence as an entity reflects the first level of self-awareness, and Kwiatkowski's self-model is not aware of the distinction between itself and the environment. In this paper, we, therefore, propose that a robot learns how to distinguish itself from the environment before acting. For this, we investigate and develop the first level of self-awareness \cite{rochat2003five} and demonstrate that a robot can get a sense of itself by simplifying the learning task to distinguishing itself in different contexts.

Hefner et al.~\cite{10.3389/fnbot.2020.00005} have reviewed biological studies and robotics studies that are related to self. The authors have concluded that the self-exploration of behaviours, body representations, and sensory-motor simulations and predictive processes are the three components that could represent self-agency in a robotic system. They have also suggested that self-agency can be measured as the prediction error. The latter is similar to what Amos et al.~\cite{DBLP:conf/iclr/AmosDCRCMETFD18} have demonstrated where they based self-awareness on predictive control models to allow a robot to create a link between itself and the environment. Similarly, Haber et al.~\cite{10.5555/3327757.3327931} has developed an intrinsically motivated agent by using world-model predictions via a supervised learning strategy to model agent awareness in order to generate different behaviours in complex environments. Similar to this work, Lanillos and Cheng \cite{7740933} have used a hierarchical Bayesian computational model to define the self in a robot and argued that the understanding of sensory mapping changes is core to self-perception. Also, Gold and Scassellati \cite{gold2007bayesian} have observed that motor actions and visual motion using probabilistic reasoning can allow a robot to self-recognise in front of a mirror and test for self-recognition of their robot's self. In this paper, we considered more environments go beyond simpler backgrounds as we include clutter within the environment in order to introduce distractors while acquiring the self. We also use a deep neural network to fuse visual and proprioception inputs to acquire a sense of self. We must note that their approaches require the robot induce motion to achieve a prediction of $81.9\%$ for self-distinction \cite{7740933}.


\section{MATERIALS \& METHODS}\label{sec:MaterialsAndMethod}

\subsection{Design and Rationale}

Our approach to artificial self-awareness focuses on building an initial sense of self in the robot by enabling it to differentiate itself (i.e. Level One in Rochat's self-awareness levels, Section \ref{sec:LR}) from the environment using proprioception and vision. For this, we design a Deep Neural Network architecture (Fig. \ref{fig:selfy_model}) to support and understand the \textit{self} in the robot. The levels of implicit self-awareness (Section \ref{sec:LR}) inspire our architecture design, and we, therefore, propose that these implicit levels can be mapped for robots as follows:

\begin{itemize}
    \item Level 0 -- "Self-Unity": We propose that this level corresponds to the robot's physical, mechanical and sensory capabilities. That is, the required sensing and motor devices and software that allow a robot to deal with the world, e.g. robot's kinematics, dynamics, sensor definition and configuration, motion planning, etc. These capabilities are interfaced via software APIs and software drivers, e.g. the Robot Operating System (ROS)\cite{noauthor_rosorg_nodate}.
    \item Level 1 -- "Differentiation": This level is the initial self, and we propose that this level is about learning how to differentiate itself by seeing its arms including arms and gripers in association with its proprioception without temporal connection between observations. The assumption at this level is that the robot has a high-level description of its limbs via forward and inverse kinematics, and can move its arms via motion planning. The objective is then to confirm if the observed arms and grippers belong to the robot.
\end{itemize}

On Level 1, the output is the initiation of self and consists of sensing the distinction between robot and environment. In Level One for humans (Section \ref{sec:LR}), \cite{rochat2003five} stated that the individual starts to think that there is a unique experience on what the person sees with respect to its proprioception. Hence, we anticipate that vision and body movement are crucial components to initiate the self in a robot. Similarly, the human brain is a prediction machine~\cite{sancaktar2019end, friston2010free}, therefore, in our approach, the prediction of \textit{self} depends mainly on two elements: the presence of the limbs in the robot's field of view and the sense of its movement.

The rationale behind our approach is to define a neural network architecture that provides a way to learn the first level of self-awareness and to understand the internal mechanisms of level one - Differentiation. The predicted output of the neural network is, therefore, a supervised binary classification task that predicts the sense of self of the robot. 

\subsection{Implementation}

\begin{figure}[t]
    \centering
    \includegraphics[width=8.5cm] {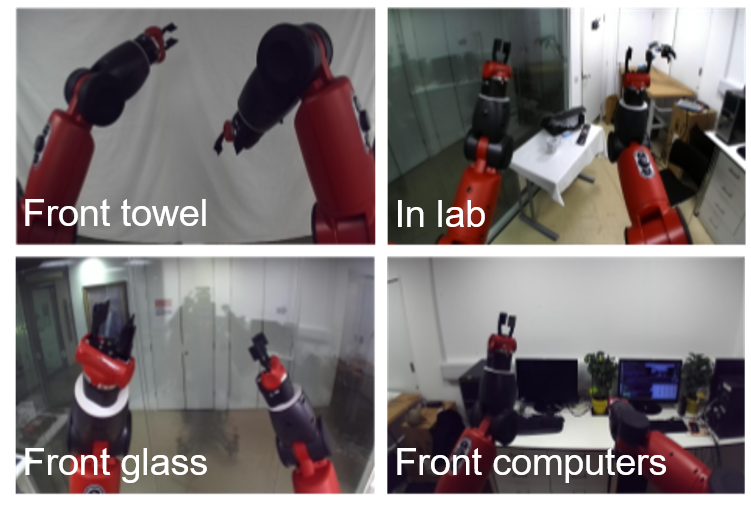}
    \caption{Sample images from captured scenes, ref. Table \ref{tab:four_groups}}
    \label{fig:mesample}
\end{figure}

To achieve Level 1, the robot uses its visual sense to discriminate its limbs together with proprioception. For this, we used the robot's vision and proprioception capabilities as the sensory inputs and combined proprioception with static visual representations to get an initial snapshot of self. Vision comprises RGB images captured using a stereo ZED camera from Stereolabs configured to output images at 720p resolution. Captured images contain a representation of the robot's arms or environment. Proprioception consists of the robot's joint states; being velocity, angular position, and motor torque.

Our architecture for Level 1 of self-awareness is shown in Fig. \ref{fig:selfy_model}) and consists of a Resnet18 network \cite{He_2016} to process the visual state of the robot. Resnet18 is a state-of-the-art architecture used widely for object detection and classification. Similarly, for proprioception, we used a single, fully connected network layer to process the internal state of the robot. The output from Resnet18 is a tensor size of $19$ that is concatenated with the output of $FC0$ proprioception tensor of size $76$. The concatenated tensor is of size $95$ and is passed to a fully connected layer - $FC1$. The output of $FC1$ is a tensor of $32$ that inputs into the fully connected layer, $FC2$, that predicts \textit{self} or \textit{environment}.

\begin{table}[tp]
\centering
\caption{\label{tab:four_groups}Experimental groups and unseen test group datasets}
\begin{tabular}{l|l|l|}
\cline{2-3}
                                       & \textbf{Experimental Groups Sets}                                                                 & \textbf{Unseen Test Group} \\ \hline
\multicolumn{1}{|l|}{\textbf{Group-1}} & \begin{tabular}[c]{@{}l@{}}\{ In lab, Front glass, \\    Front towel \}\end{tabular}         & Front computers            \\ \hline
\multicolumn{1}{|l|}{\textbf{Group-2}} & \begin{tabular}[c]{@{}l@{}}\{ Front computers, In lab, \\    Front glass \}\end{tabular}     & Front towel                \\ \hline
\multicolumn{1}{|l|}{\textbf{Group-3}} & \begin{tabular}[c]{@{}l@{}}\{ Front computers, Front glass,\\    Front towel \}\end{tabular} & In lab                     \\ \hline
\multicolumn{1}{|l|}{\textbf{Group-4}} & \begin{tabular}[c]{@{}l@{}}\{ Front computers, In lab,\\  Front towel \}\end{tabular}          & Front glass                \\ \hline
\end{tabular}
\end{table}

\begin{figure*}[ht]
    \centering
    \includegraphics[width=15cm]{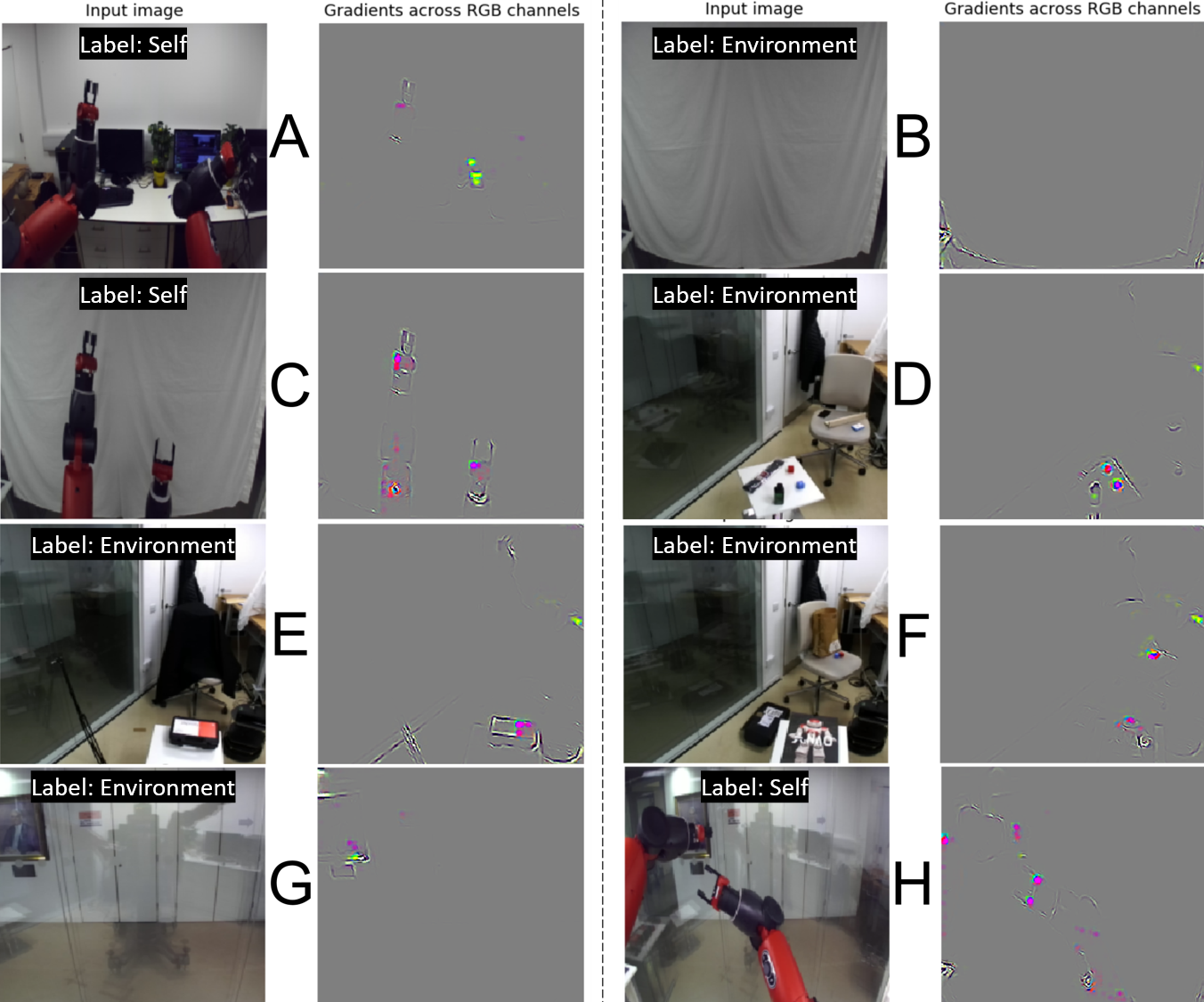}
    \caption{This images are representing the saliency maps of different environment groups as described in Table \ref{tab:four_groups}, were A corresponds to Group-1; B and C, to Group-2; D and F, to Group-3; and, G and H, to Group-4. For each group, the right image shows the predicted label, and the left images shows the regions the model focused on.}
    \label{fig:saliencymap}
\end{figure*}

We implemented a ROS node to capture and store synchronised visual and proprioceptive sensor information. During data capturing, we recorded Baxter moving its hands using random gripper poses for both arms and within its working volume. A total of $30$k images and proprioception states were captured over four different environmental settings, as shown in Fig. \ref{fig:mesample}. Each scene represents a unique group that ranges from simple (front towel and front glass) to cluttered (\textit{front computers} and \textit{in lab}) environments. Each experimental group includes two classification labels, namely \textit{self} and \textit{environment}.

An experimental group is a combination of three scenes while leaving one out for testing purposes, as shown in Table \ref{tab:four_groups}. The objective is to have broader and diverse data groups for training our proposed architecture (Fig. \ref{fig:selfy_model}). We split 80/20 proportions each experimental group for training and validation, respectively~\cite{guyon1997scaling}. Accordingly, the training and validation sets represent about $20$k and $5$k images and proprioception states, respectively. We used PyTorch to implement our level 1 of artificial self-awareness architecture with PyTorch's default cross-entropy loss. We used a pre-trained version of Resnet18 and fine-tuned it with our dataset. Training consisted of $24$ epochs each with a $64$ batch and a learning rate of $0.001$.


\section{EXPERIMENTS}\label{sec:level1exp}

\section{RESULTS \& DISCUSSION}\label{sec:rsanddis}

\begin{figure*}[t]
    \centering
    \includegraphics[width=14cm]{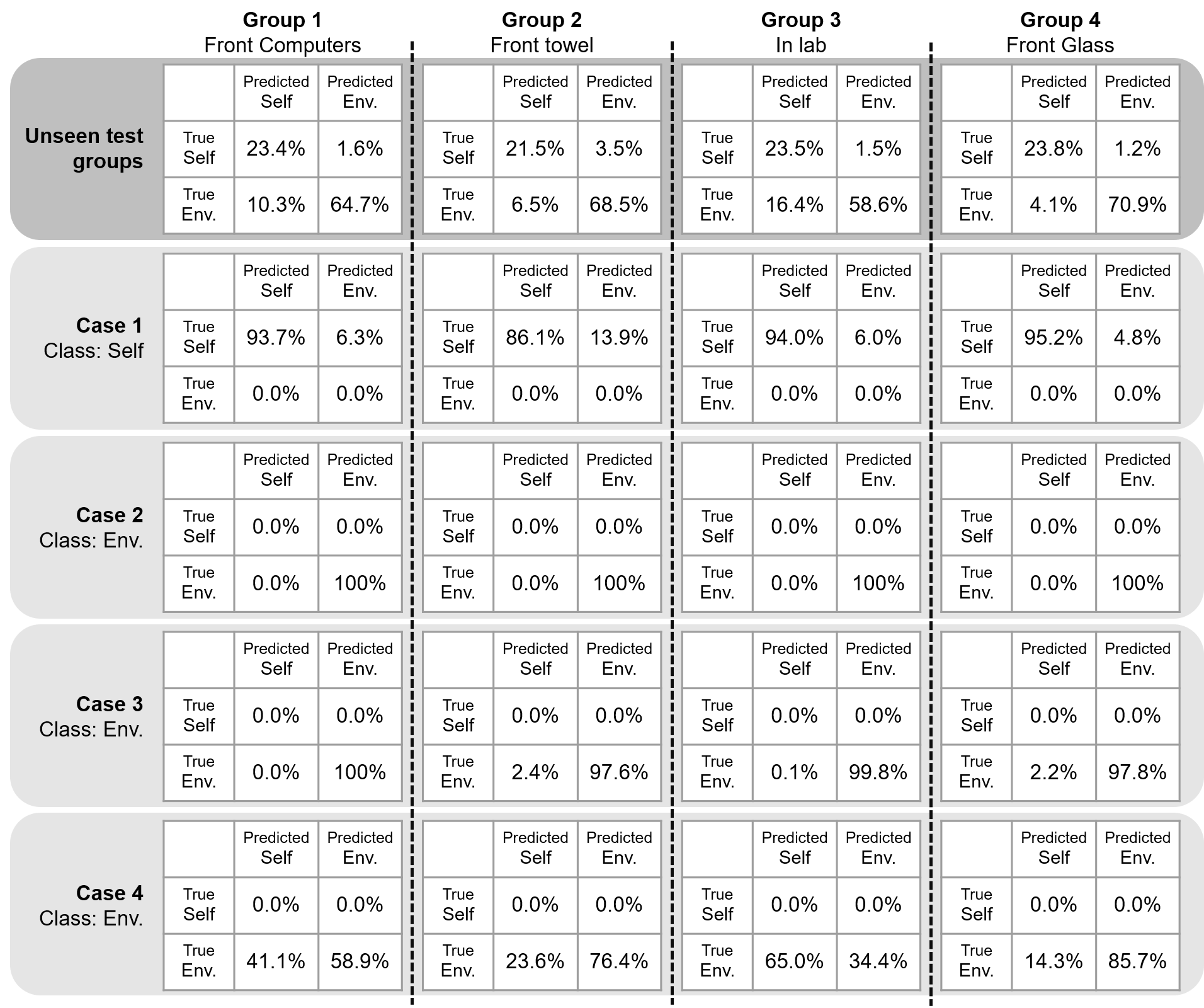}
    \caption{The first row shows the unseen test groups classification accuracies, while the next four rows show the experimental cases for each unseen group.}
    \label{tbl:result_groups_and_cases}
\end{figure*}

The robot's capabilities frame our experimental design. That is, the robot cannot move to a different place in the room. Similarly, the robot's vision sensor is fixed on top of the robot's head and cannot actively move its head. The robot can move its arms freely within its predefined working volume, and there are no obstacles included in each of the experimental groups in Table \ref{tab:four_groups}. According to Rochat~\cite{rochat2003five}, newborns wave their hands randomly in order to try to identify the objects in front of them. We, therefore, allowed the robot to wave its limbs without a predefined task in the environment to enable the robot to learn to perceive and differentiate itself from the environment.

To further test our hypothesis, we carry out an ablation study represented by four confounding cases to understand the effectiveness of the combination of the proprioception and vision within our proposed model. This experiment, therefore, consists of four confounding experimental cases (Table \ref{tab:4cases}) that compare unseen experimental groups against confounding scenarios the robot may encounter. The objective is to confirm that the robot can differentiate itself with a certain degree of certainty while presented with confounding sensor signals. Thus, \textit{Case-1} comprises the unseen test group where images and proprioception corresponds to the \textit{self} class; while \textit{Case-2}, images and proprioception correspond to the \textit{environment} class. \textit{Case-3} comprises confounding samples where the robot's arms are in the visual field of the robot, but the robot's proprioception corresponds to the \textit{environment} class. While \textit{Case-4} is composed of environment images but the robot's proprioception comes from the \textit{self} class.

To understand Level 1, we adopted a leave-one-out cross-validation strategy to test each trained experimental group (Table \ref{tab:four_groups}). By having an unseen experimental group, we are then able to verify the validity of our hypothesis that \textit{Level 1 for artificial self-awareness in the robot increases its self-certainty in an unseen environment}. Accordingly, confusion matrices for each unseen test group in Table \ref{tab:four_groups} are shown in Fig. \ref{tbl:result_groups_and_cases}. The classification accuracy for each unseen test group is: Group-1 is $88.1\%$; Group-2, $90\%$, Group-3, $82.1\%$, and Group-4, $94.7\%$. We can, therefore, state that our architecture enables the robot to differentiate itself from the environment with an average certainty of $88.7\%$.

We have used FlashTorch~\cite{misa_ogura_2019_3461737} to gain insights on how our Level 1 approach perceives images by means of a saliency map, as shown in Fig. \ref{fig:saliencymap}. A saliency map is an image showing which pixel regions the neural network focused on in order to predict the underlying output. Accordingly, FlashTorch allows us to gain insights on which regions in the image contributed the most to predict \textit{self} or \textit{environment}. For this, vision and proprioception are used to predict the class, but we only evaluate the vision component of our architecture with FlashTorch.

As observed in Fig. \ref{fig:saliencymap}, our Level 1 architecture is biased towards bright colours and distinctive features that remained constant in the robot's visual field. We must note that clutter is semi-consistent in the scene during data capturing, i.e. clutter was randomly placed. Clutter resulted in making our architecture to focus on objects that did not observe significant variation in their pose. For example, in the \textit{Front Computers} test group (ref. Table \ref{tab:four_groups}), $10.3\%$ of the \textit{environment} class is classified as \textit{self} (Fig. \ref{tbl:result_groups_and_cases}, unseen test groups row). The reason is that the network interpreted the yellow pot in Fig. \ref{fig:saliencymap}-A as part of the robot (i.e. \textit{self} class). Close inspection of Fig. \ref{fig:saliencymap}-B and -C (\textit{front towel group}) shows that when the robot's hands are within the field of view, environment features such as the edges of the towel do not contribute to predicting the correct class, observing $90\%$ accuracy for predicting the correct class. Figures \ref{fig:saliencymap}-D, -E, and -F (\textit{in lab} unseen test group) reveal the sensitivity of the network towards similar colours to the robot used in our experiments (i.e. red) and bright colours. The latter represents $17.9\%$ of incorrect classifications. The \textit{front glass} unseen test group (Fig. \ref{fig:saliencymap}-G and -H) achieves a $94.7\%$ of classification accuracy; however, our Level 1 architecture is biased towards bright regions in the images (i.e. picture frame in the background and table corner in the left-bottom).

\begin{table}[t]
\caption{\label{tab:4cases}Confounding experimental cases}
\begin{tabular}{l|l|l|}
\cline{2-3}
\multicolumn{1}{c|}{} & \multicolumn{1}{c|}{\textbf{Class}} & \multicolumn{1}{c|}{\textbf{Description}} \\ \hline
\multicolumn{1}{|l|}{\textbf{Case-1}} & \multicolumn{1}{c|}{Self} & \begin{tabular}[c]{@{}l@{}}Vision and proprioception correspond to the\\ robot's arms being in the field of view\end{tabular}                                \\ \hline
\multicolumn{1}{|l|}{\textbf{Case-2}} & \multicolumn{1}{c|}{Environment} & \begin{tabular}[c]{@{}l@{}}Vision and proprioception correspond to the\\ robot's arms not being in the field of view\end{tabular}                          \\ \hline
\multicolumn{1}{|l|}{\textbf{Case-3}} & Environment                      & \begin{tabular}[c]{@{}l@{}}The robot's arms are in the field of view\\ but the proprioception matches the\\ environment class\end{tabular} \\ \hline
\multicolumn{1}{|l|}{\textbf{Case-4}} & Environment                      & \begin{tabular}[c]{@{}l@{}}Proprioception corresponds to the self\\ class but the robot's arms are not in the\\ field of view\end{tabular}     \\ \hline
\end{tabular}
\end{table}

\begin{figure}[t]
    \centering
    \includegraphics[width=8.5cm]{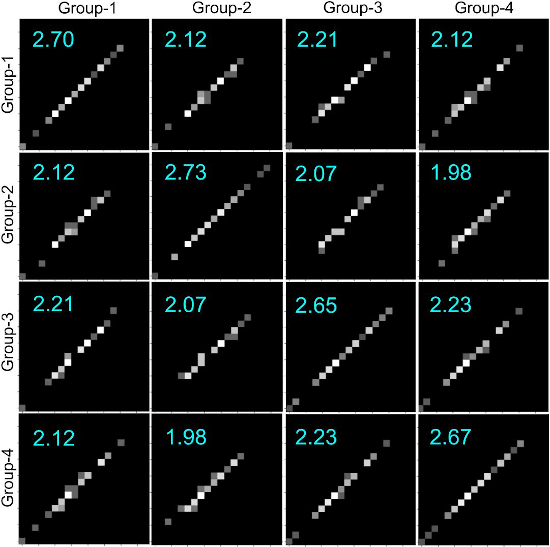}
    \caption{Mutual information and joint 2D histograms of the trained weights for four Level 1 architectures. The mutual information is noted at the top left corner on each joint histogram plot.}
    \label{fig:mutual_histogram}
\end{figure}

The unseen test Group-1 (\textit{front computers}) and Group-3 (\textit{in lab}) have noticeable classification errors of $11.9\%$ and $17.9\%$, respectively. To investigate these errors, we split the unseen test set into four confounding cases, as described in Table \ref{tab:4cases}. These results are shown in Table \ref{tbl:result_groups_and_cases}, and reveal that Case-4, is the most misclassified case in both Group-1 and Group-3. The reason for these misclassifications is that Resnet18 is biased towards bright colours as discussed above. If we compare with Group-2 and Group-4 where the robot is facing uncluttered environments, Case-4 yields higher classification scores since visual false positives are kept at a minimum. That is, we can observe in Fig. \ref{fig:saliencymap}-C and -B that when the robot's hands become more predominant on uncluttered backgrounds, our Level 1 architecture predicts the correct classification regardless of confounding signals coming from proprioception. We also noticed in Fig. \ref{tbl:result_groups_and_cases} that proprioception signals have a high contribution to predicting the correct class. For instance, in case-3 where images contain the robot's arms, but proprioception corresponds to the \textit{environment} class, our architecture can predict the correct class with high accuracy for all groups.

To further understand whether our Level 1 architecture learns to differentiate the robot from the environment, we computed the Mutual Information \cite{fang2018dissecting} for each group's train dataset (Table \ref{tab:four_groups}). Our objective is to measure and compare if four Level 1 trained architectures have a degree of similar knowledge that it is invariant to the training set. Mutual information allows us to compare multimodal sources and measure how well two sources are matched by mutual dependence between two variables. That is, different sources of information means more distributed points in the joint histogram and, consequently, low mutual information metric.

The spread in the joint histogram is associated with uncertainty, and in Fig.~\ref{fig:mutual_histogram}, joint histograms show minor variability in the correlation between the group's models weights. The latter shows that there are no significant differences between the trained models despite the differences in the training datasets, and the misclassification in the confusion matrices results (Table \ref{tab:four_groups}) are based on the environment noise as other objects within the environment distract the network attention. Since mutual information is computed at the last layer of our architecture, proprioception is taken into account during the classification. Therefore, this demonstrates that our Level 1 network architecture captures a degree of self-awareness and, consequently, certainty. We can, therefore, conclude our experimental hypothesis in Section \ref{sec:level1exp} holds for the experiments presented in this paper. In the companion video to this paper, we demonstrate our approach to Level 1 artificial self-awareness. That is, we use a Baxter robot to predict the sense of self while observing what is in front of it, i.e. self or environment.

\section{CONCLUSIONS \& FUTURE WORK}\label{sec:concAndFuture}

In this paper, we presented an approach to Level 1 of artificial self-awareness in a dual-arm robot. Our approach is inspired by the first level of self-awareness defined by Rochat \cite{rochat2003five}. By using vision and proprioception, we have demonstrated that a robot can differentiate itself from the environment with an average classification accuracy of $88.7\%$ using unseen test samples and across four different scenes' groups presented in Fig. \ref{tbl:result_groups_and_cases}.

An initial self is defined in the robot, but the robot cannot locate its limbs within the environment and put them into context for a task. Thus, future work comprises developing Level 2 (Situation; Section \ref{sec:LR}) of artificial self-awareness. The idea is to employ temporal sequences of the robot's arms, and model visual and proprioception experiences in terms of a recurrent network architecture, which we believe is the next step to let a robot to be able to identify itself with higher self-certainty in an environment.

\section*{ACKNOWLEDGMENT}
Ali AlQallaf thanks the Kuwait Institute for Scientific Research. We also thank the support of NVIDIA Corporation for the donation of the Titan Xp GPU used in this research.

\bibliographystyle{IEEEtran}
\bibliography{bibliography.bib}

\end{document}